\documentclass[conference]{IEEEtran}
\IEEEoverridecommandlockouts
\usepackage{cite}
\usepackage{amsmath,amssymb,amsfonts}
\usepackage{algorithmic}
\usepackage{graphicx}
\usepackage{textcomp}
\usepackage{xcolor}
\usepackage[colorlinks]{hyperref}
\def\BibTeX{{\rm B\kern-.05em{\sc i\kern-.025em b}\kern-.08em
    T\kern-.1667em\lower.7ex\hbox{E}\kern-.125emX}}

\usepackage{multirow}
\usepackage{bm}
\usepackage{amssymb}
\usepackage{pifont}
\newcommand{\cmark}{\ding{51}}%
\newcommand{\xmark}{\ding{55}}%

\begin{document}
\title{On the Evaluation of Prohibited Item Classification and Detection in Volumetric 3D Computed Tomography Baggage Security Screening Imagery
}

\author{\IEEEauthorblockN{Qian Wang}
\IEEEauthorblockA{\textit{Department of Computer Science} \\
\textit{Durham University}\\
Durham, UK \\
}
\and
\IEEEauthorblockN{Neelanjan Bhowmik}
\IEEEauthorblockA{\textit{Department of Computer Science} \\
\textit{Durham University}\\
Durham, UK \\
}
\and
\IEEEauthorblockN{Toby P. Breckon}
\IEEEauthorblockA{\textit{Department of \{Computer Science $|$ Engineering\}} \\
\textit{Durham University}\\
Durham, UK \\
}
}

\maketitle
\begin{abstract}
X-ray Computed Tomography (CT) based 3D imaging is widely used in airports for aviation security screening whilst prior work on prohibited item detection focuses primarily on 2D X-ray imagery. In this paper, we aim to evaluate the possibility of extending the automatic prohibited item detection from 2D X-ray imagery to volumetric 3D CT baggage security screening imagery. To these ends, we take advantage of 3D Convolutional Neural Neworks (CNN) and popular object detection frameworks such as RetinaNet and Faster R-CNN in our work. As the first attempt to use 3D CNN for volumetric 3D CT baggage security screening, we first evaluate different CNN architectures on the classification of isolated prohibited item volumes and compare against traditional methods which use hand-crafted features. Subsequently, we evaluate object detection performance of different architectures on volumetric 3D CT baggage images. The results of our experiments on \textit{Bottle} and \textit{Handgun} datasets demonstrate that 3D CNN models can achieve comparable performance ($\sim$98\% true positive rate and $\sim$1.5\% false positive rate) to traditional methods but require significantly less time for inference (0.014s per volume). Furthermore, the extended 3D object detection models achieve promising performance in detecting prohibited items within volumetric 3D CT baggage imagery with $\sim$76\% mAP for bottles and $\sim$88\% mAP for handguns, which shows both the challenge and promise of such threat detection within 3D CT X-ray security imagery. 
\end{abstract}


\section{Introduction}
X-ray baggage security screening is widely used to maintain aviation security. Currently, multi-view X-ray is predominantly used in aviation security for cabin baggage screening. This traditional baggage screening process, using 2D X-ray scanners, has the disadvantage of both inter-object occlusion and clutter within any given image projection of the scanned baggage item. As a result, it poses a considerably challenging visual search task for the human operators to discover the prohibited items (e.g., liquids, firearms, knives, etc.) overlapped with other benign items (e.g., electronic devices) within a constrained time frame. For this reason, passengers are currently required to divest large electronic devices and liquids which decreases checkpoint throughput significantly. Furthermore, human operator performance can be subjective and is heavily affected by many factors such as the experience, fatigue, monotony and concentration, although many successful measures have been taken to alleviate the problem in practice (e.g., Threat Image Projection (TIP) \cite{bhowmik2019good,wang2020reference} and shorter shift rotations \cite{meuter2016and}).

By leveraging recent advances in object classification and detection, significant progress has been made in automatic prohibited item detection within 2D X-ray imagery \cite{akcay2018using}. The use of deep learning techniques allows real-time and accurate detection of prohibited items even in cluttered X-ray images \cite{gaus2019evaluating,bhowmik19subcomponent,bhowmik19electronics}. However, performance can be affected when the baggage contains significant clutter and inter-object occlusion due to the fundamental limitation of projected 2D X-ray imagery.
To improve the detection rate without affecting the checkpoint throughput, airports are currently increasing the use of 3D CT screening which does not require the removal of electronic devices and liquids during baggage screening. The reconstructed 3D CT images provide more information and make it possible for the human operators to inspect the 3D CT images from differing views. However, current technology does not facilitate the automatic detection of (non-explosive) prohibited items such as weapons and liquid containers. It is unknown \textit{if the success of deep learning approaches in 2D X-ray imagery can be similarly replicated in volumetric 3D CT imagery for baggage security screening} and \textit{whether the 3D CNN based approaches are efficient enough for operational viability?}

To answer the above questions, in this paper we extend the prohibited item classification and detection methods from 2D to 3D imagery and evaluate their effectiveness in real volumetric 3D CT baggage security screening imagery. Firstly, we look into the task of 3D object classification for isolated prohibited items in volumetric 3D CT data. We investigate different CNN architectures including ResNet \cite{he2016deep} with variable depths and Voxception-ResNet \cite{brock2016generative}. We also evaluate the effectiveness of data and feature augmentation techniques in 3D CNN based classification. As for the detection problem, we consider two successful object detection frameworks for 2D imagery: Faster R-CNN \cite{ren2015fasterrcnn} and RetinaNet \cite{lin2017focal}.  

The contributions of this work are summarized as follows:
\begin{itemize}
    \item[--] the first attempt to use deep CNN models for the prohibited item classification and detection within volumetric 3D CT baggage imagery to our best knowledge;
    \item[--] an evaluation of different 3D CNN models in the classification of prohibited items within volumetric 3D CT baggage imagery and the effect of data/feature augmentation;
    \item[--] an evaluation of prohibited item detection within volumetric 3D CT baggage imagery using 3D Faster R-CNN and 3D RetinaNet CNN architectures.
\end{itemize}

\section{Related Work}\label{sec:related}
The work presented in this paper is closely related to some prior art in two aspects which we briefly discuss in this section: \textit{3D baggage imagery analysis} and \textit{3D CNN}. 
\subsection{3D Baggage Imagery Analysis}
To enable automatic baggage screening using 3D CT imagery, a variety of studies have been carried out in recent years \cite{wiley2012automatic,flitton20123d,mouton20143d,jin2015joint,flitton2015object,mouton2015materials,mouton2015review,wang2019approach,wang2020reference}.

One research direction is object segmentation based on the material and morphological structure \cite{wiley2012automatic,mouton2015materials,wang2019approach}. Specifically, Mouton et al. \cite{mouton2015materials} proposed a two-stage approach for object segmentation within 3D CT imagery. A CT volume is firstly coarsely segmented based on the voxel intensity ranges of pre-defined materials. Subsequently, a variety of shape descriptors are computed as features for the random forest classifier to determine a segment resulted from the first stage is good (containing only one object) or bad (containing multiple objects and hence need further segmentation).
Wang et al. \cite{wang2019approach} studied the issue of object segmentation and classification in 3D CT imagery and focused mainly on the material characteristics without considering any specific prohibited item (e.g., firearm, knife, etc.). An approach to 3D segmentation was proposed based on recursive morphological operations and the Support Vector Machines (SVM) were employed for the classification of three types of materials.

3D object classification was studied in \cite{flitton20123d,mouton20143d,flitton2015object} where a binary classifier was formulated to distinguish the objects of interest (i.e. handgun or bottle) from the background volumes which contain cluttered content (e.g., books, clothes and etc.). The bag-of-word features and a SVM classifier were used in these studies (denoted Cortex \cite{flitton20123d}, Codebook \cite{flitton2015object} and ERC \cite{mouton20143d} in Table \ref{Table:comp}). Isolated 3D CT volumes of prohibited items are manually cropped from the baggage CT images to form the positive sample set which are also employed here in our work for the evaluation of 3D CNN based classification. 

More recently, Wang et al. \cite{wang2020reference} present an approach to 3D threat image projection which can be used to generate realistic and plausible volumetric 3D CT baggage images with superimposed threat object signatures. This technique can be used for training not only human operators for baggage security screening but also machine learning algorithms for automatic prohibited item detection without time-consuming data collection and manual annotation such as in \cite{bhowmik2019good}.

\subsection{3D Convolutional Neural Networks}
3D CNN models are widely used for object classification and detection within varying data modalities such as LiDAR point cloud \cite{maturana2015voxnet,zhou2018voxelnet}, RGB-Depth data \cite{qi2018frustum}, 3D Computer Aided Design (CAD) models \cite{brock2016generative} and medical CT imagery\cite{xie2019automated,jaeger2019retina}. 

VoxelNet \cite{zhou2018voxelnet} is an end-to-end 3D object detector specially designed for LiDAR data. It consists of three modules: feature learning network (subdivide the point cloud into many subvolumes/voxels, feature engineering + fully connected neural network), convolutional middle layer (3D convolution applied to the stacked voxel feature volumes, each subvolume/voxel is a feature vector) and region proposal networks.
VoxNet \cite{maturana2015voxnet} in a more generic model being able to handle different types of 3D data including LiDAR point cloud, CAD and RGBD data.
Qi et al. \cite{qi2016volumetric} improved the performance of VoxNet by introducing the auxiliary subvolume supervision to alleviate the overfitting issue.

RGB-Depth data can also be processed using 3D CNN by firstly extracting proposals from 2D RGB images using a 2D object detector and transforming the proposals and corresponding depth information into 3D point clouds \cite{qi2018frustum}. The generated 3D point clouds can be further explored by 3D CNN models such as PointNet \cite{qi2017pointnet}.

These models designed for point clouds, RGBD data, CAD models or medical CT images are not readily transferable to our volumetric 3D CT imagery for baggage security screening since the modality of input data for 3D CNN can differ significantly. However, the design of 3D CNN architectures and the training strategies used in existing work can be repurposed towards our prohibited item classification and detection within 3D CT baggage imagery.
\section{Method} \label{sec:method}
In this section, we describe the methods used in our work for prohibited item classification and detection within volumetric 3D CT baggage imagery. We firstly consider a classification problem to evaluate the possibility of discriminating the isolated prohibited item signatures from benign CT volumes. Subsequently, we consider the more realistic detection problem which aims to not only classify the prohibited item within a baggage CT image but also localises it by generating a 3D bounding box around the target object.
\subsection{3D Prohibited Item Classification}\label{sec:method_cls}
Our prohibited item classification is formulated as a binary classification problem in this work to evaluate the effectiveness of 3D CNN models. Specifically, given a CT volume as the input, the 3D CNN model aims to determine if the volume contains a prohibited item signature (positive sample) or not (negative sample). The positive samples are manually cropped from baggage volumes, hence they can have varying dimensions and orientations. As a result, we need to pre-process the input samples to a common voxel scaling before feeding them into the 3D CNN.
\subsubsection{3D CNN Model} \label{sec:3dcnn}
\begin{figure}
    \centering
    {\includegraphics[width=0.45\textwidth]{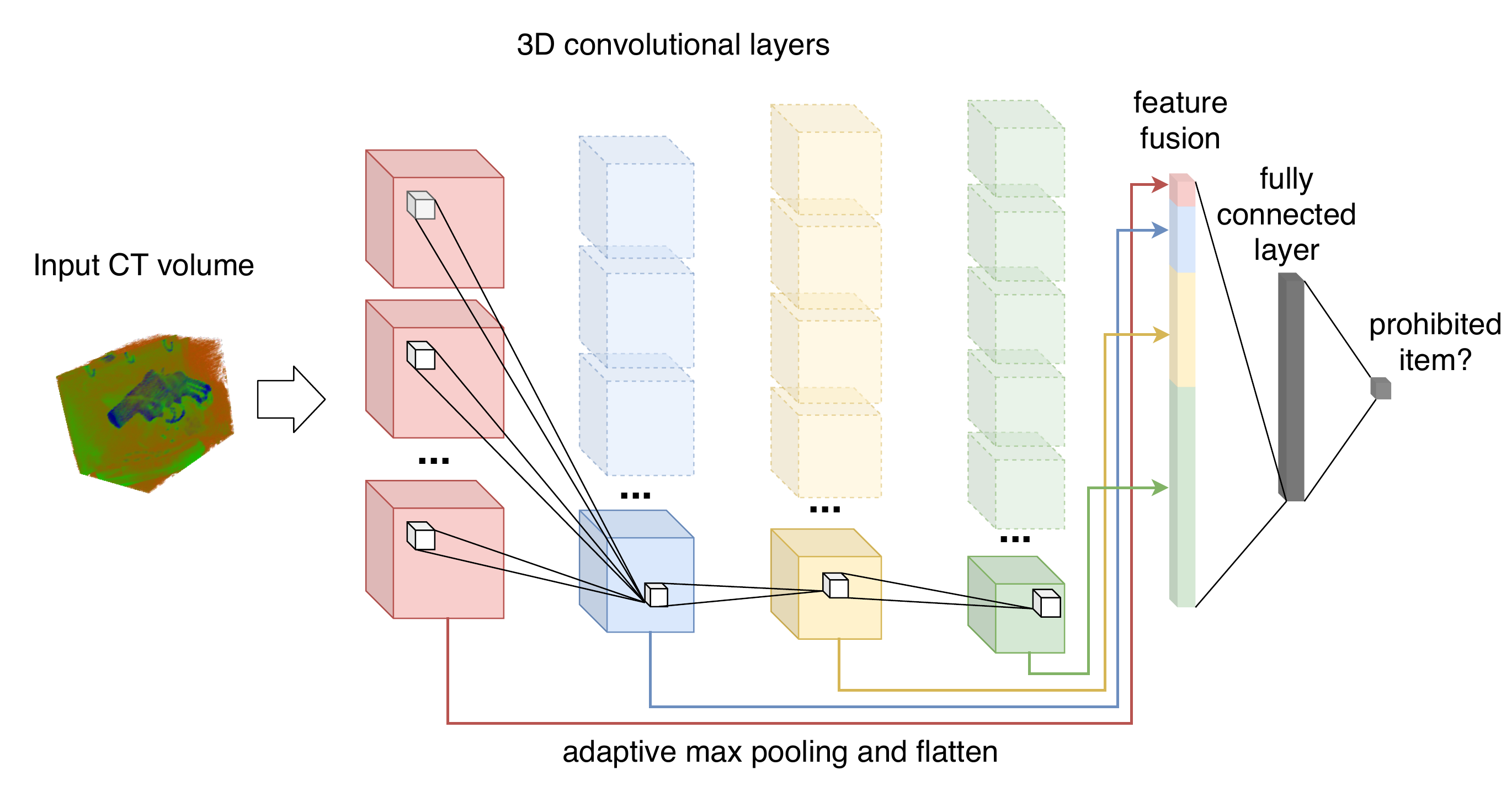}}
    {\caption{Our 3D CNN architecture for classification with rich feature fusion based on ResNet \cite{he2016deep} (four CNN layers used for feature fusion are depicted here whilst many other intermediate layers are omitted; the output layer has one node for our binary classification problems). }
        \label{fig:3dcnn}}
\end{figure}

We consider ResNet \cite{he2016deep} and Voxception-ResNet (VRN) \cite{brock2016generative} in our evaluation. ResNet was extended to the 3D version by Chen et al. \cite{chen2019med3d} for medical CT image analysis. To address the issue of variable sizes of prohibited items, we employ the idea of rich features \cite{girshick2014rich} to explore the multi-scale feature volumes. Specifically, we augment the features for fully-connected layers by fusing the feature volumes generated by multiple intermediate convolutional layers. The proposed architecture of \textit{rich feature} ResNet is illustrated in Figure \ref{fig:3dcnn}. The architecture is composed of four sequential blocks, each of which contains multiple 3D convolutional layers. By stacking different numbers of layers, we investigate variants of ResNet (ResNet$_{10}$, ResNet$_{18}$ and ResNet$_{34}$) in our experiments. Even deeper ResNet models (e.g., ResNet$_{50}$ and ResNet$_{101}$) are also investigated. However, these deeper models suffer with convergence problems.

Alternatively, we also consider a variation of deeper ResNet: Voxception-ResNet (VRN). VRN is designed by combining the ideas of Inception-style \cite{szegedy2016rethinking} networks and ResNet in a 3D CNN framework. It takes advantage of the Inception-style architectures for multi-scale visual information exploration and the advantage of residual connections for efficient training.

\subsubsection{Data Pre-processing and Augmentation}\label{sec:preprocess}
Correctly designed data pre-processing and augmentation strategies are beneficial for training CNN models for small datasets. Here we consider two strategies for data pre-processing and augmentation: \textit{rescaling} and \textit{rotation}. Since our chosen CNN architecture uses an adaptive pooling layer before the fully-connected layers to handle the variable dimensions of the feature volumes caused by the varying input sizes, the input volumes are not required to have the same size. The \textit{rescaling} aims to restrict three dimensions (i.e. height, width and depth) of input volumes within a limited range. Specifically, we rescale the input 3D volumes to have dimensions no greater than a pre-defined number of voxels in any dimension by down-sampling a given volume $\bm{V}\in \mathbb{R}^{H\times W \times D}$ by factors of $max\{1,\lfloor H/s \rfloor \}$, $max\{1,\lfloor W/s \rfloor \}$ and $max\{1,\lfloor D/s \rfloor \}$ for all three dimensions respectively. The hyper-parameter scaling value $s$ is empirically chosen as 32 for favourable classification performance.

During training, the 3D input volumes are randomly rotated in three planes (i.e., $xy$, $yz$ and $xz$) with a fixed probability to augment the training data. The augmentation of \textit{rotation} is enabled randomly in one of the three planes and the rotation angles are restricted to $\{\it{90, 180, 270}\}$ degrees. Without this restriction  the volumes after rotation become more complicated by the requirement for resampling and zero padding and this poses an additional challenge for training.

\subsection{3D Prohibited Item Detection}\label{sec:method_det}
Prohibited item detection within volumetric 3D CT baggage imagery is a more challenging problem aiming to both localise and classify the prohibited items simultaneously. We look into the possibility of extending 2D object detection frameworks to resolve this problem arising from real-world applications of aviation security. Faster R-CNN \cite{ren2015fasterrcnn} and RetinaNet \cite{lin2017focal} are considered for their superior performance in 2D object detection within this domain \cite{gaus2019evaluating,bhowmik2019good}.

Faster R-CNN consists of three modules: Feature Extraction Network, Region Proposal Network and Region of Interest (RoI) pooling. We use ResNet$_{50}$ and ResNet$_{101}$ as the backbone networks for feature extraction. To handle the object scale variability, we use a Feature Pyramid Network similar to the rich feature extraction strategy used in the classification models. 
By contrast, RetinaNet is a one-stage object detection approach. Again, ResNet$_{50}$ and ResNet$_{101}$ are used as the backbone networks for feature extraction. For both methods the anchors are defined on multiple levels of feature volumes.

\begin{figure}
    \centering
    {\includegraphics[width=0.4\textwidth]{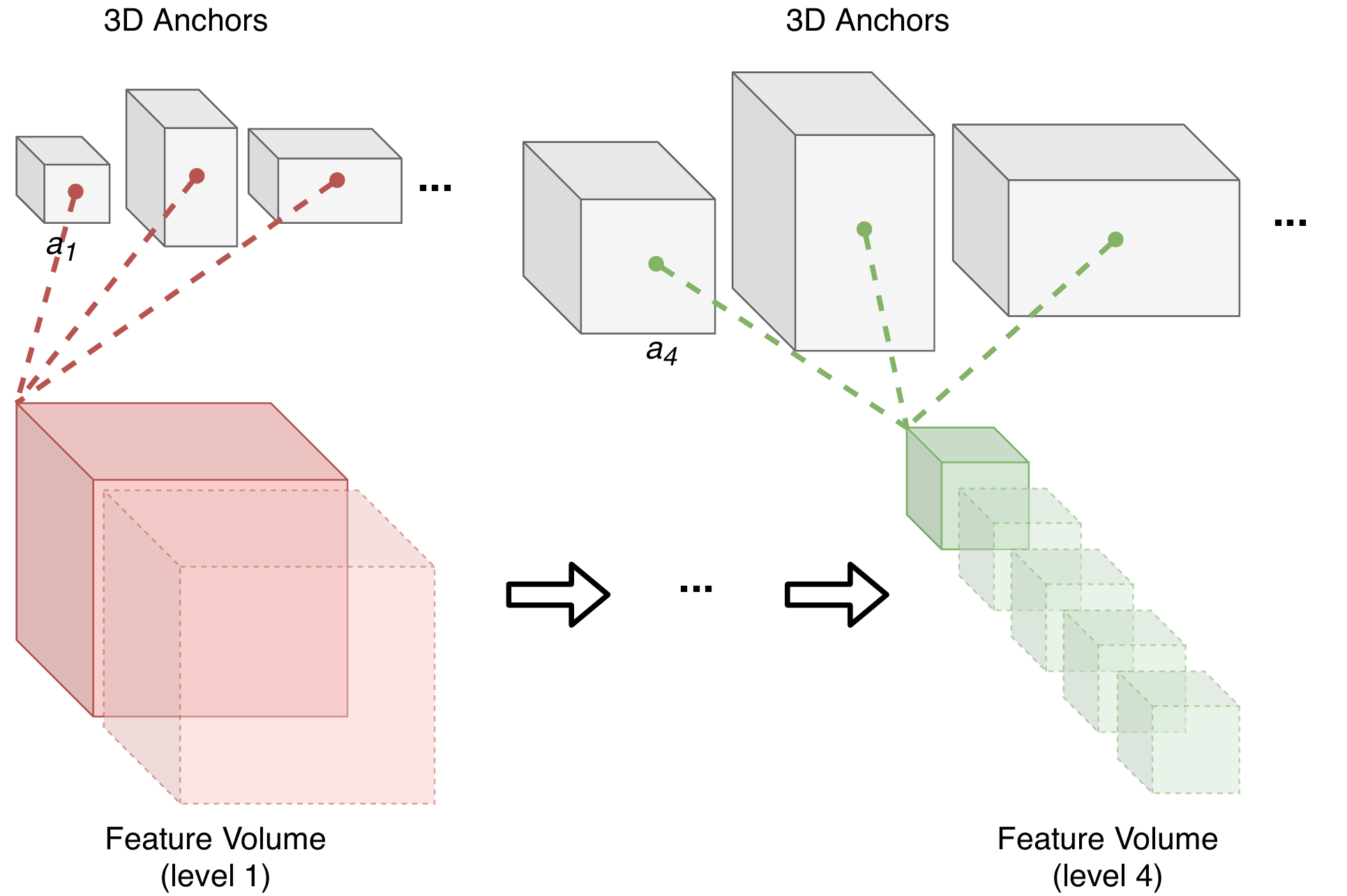}}
    {\caption{An illustration of 3D anchors built on multi-level feature volumes (small anchors are more densely built with references to all voxels within a lower-level 3D feature volume). }
        \label{fig:3danchor}}
\end{figure}

The anchor size is an important factor affecting the performance of detectors used in our evaluation. We follow the work on 2D object detection \cite{ren2015fasterrcnn,lin2017focal} and extend it to our 3D object detection frameworks. We use a set of anchor sizes and ratios to generate diverse 3D anchors in four feature pyramid levels. Specifically, we set anchor sizes as $\{a_l\}, l=1,2,3,4$ for four feature pyramid levels respectively. The anchor ratios are uniformly set as (\textit{height:width:depth}) $\{1:2:\sqrt{2}; 1:1:1; 2:1:\sqrt{2}\}$. These ratios are empirically selected rather than generated by $k$-means\cite{redmon2016you} clustering over training bounding boxes for the fact that the prohibited items within baggage can have arbitrary orientations leading to arbitrary box ratios even the items themselves have fixed dimension ratios. With the combination of anchor sizes and ratios, three 3D anchors are generated for each voxel in the feature volume for Faster R-CNN. In the RetinaNet framework, the anchors are augmented by adding extra anchor sizes of $\{a_l 2^{1/3}, a_l 2^{2/3}\}$ for all feature pyramid levels $l=1,2,3,4$ \cite{lin2017focal}. As a result, for each voxel in a feature volume nine 3D anchors are generated for RetinaNet. The anchor sizes of lower-level feature volumes should be smaller since the anchors are more densely built as shown in Figure \ref{fig:3danchor}. The effect of varying anchor sizes will be evaluated in our experiments.

\section{Experimental Setup}\label{sec:experiments}
In this section, we describe the experimental setup for the evaluation of prohibited item classification and detection within baggage CT volumes. We describe the datasets used in our experiments and implementation details of the classification and detection methods.
\subsection{Dataset}
We use the same datasets as employed in \cite{flitton2015object} for classification and detection tasks. The data was obtained from a CT80-DR dual-energy baggage-CT scanner manufactured by Reveal Imaging Inc. Two object categories (i.e. bottles and handguns) are considered in our experiments for proof-of-concept. 

\begin{figure*}
    \centering
    {\includegraphics[width=0.93\textwidth]{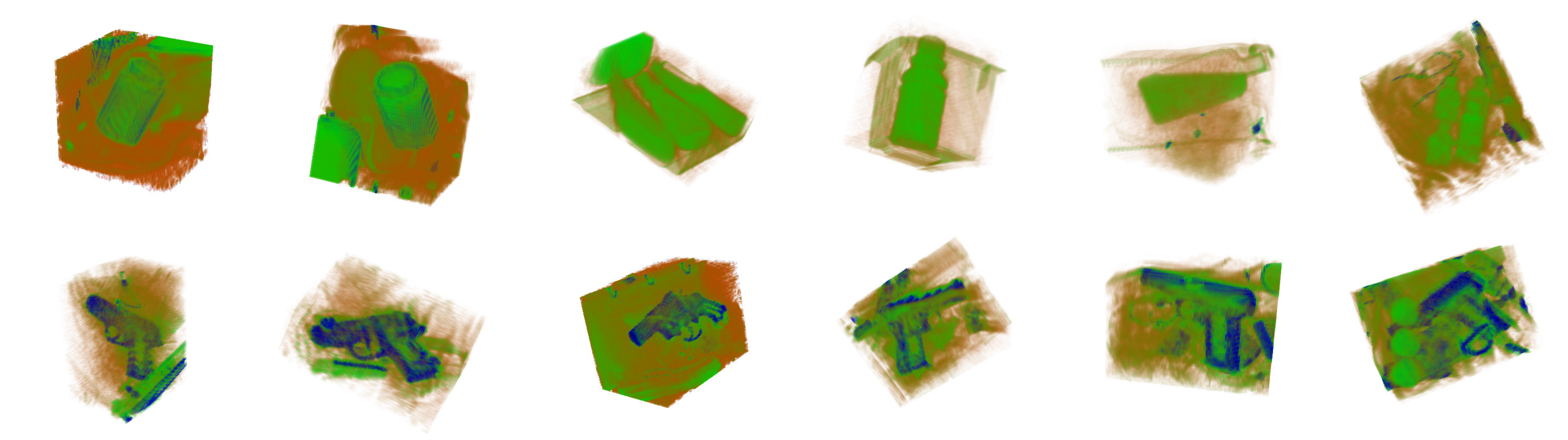}}
    {\caption{Exemplar CT volumes of isolated prohibited items (bottles in the upper row and handguns in the bottom row) used in our classification experiments. }
        \label{fig:datasample}}
\end{figure*}

For classification, we use the manually isolated CT volumes from the original baggage CT volumes. These isolated CT volumes form two independent datasets. The \textit{Bottle} dataset contains 1704 isolated CT volumes among which 526 are positive samples (i.e. with bottles) and 1178 are negative samples (i.e. without bottles). The \textit{Handgun} dataset contains 1255 isolated CT volumes among which there are 284 positive samples and 971 negative ones. Some exemplar isolated CT volumes of bottles and handguns are shown in Figure \ref{fig:datasample}. The same ten-fold cross-validation used in \cite{flitton2015object} was employed in our experiments for classification.

For detection, we use the original whole baggage CT volumes. Again two datasets (i.e. \textit{Bottle} and \textit{Handgun}) are considered independently in our experiments. There are 305 baggage volumes in the \textit{Bottle} dataset and 526 bottle signatures are annotated by 3D bounding boxes. There are 267 baggage volumes in the \textit{Handgun} dataset within which 282 handgun signatures are annotated by 3D bounding boxes. We divide the dataset into training (80\%) and test (20\%) subsets randomly. Three random splits are generated for each dataset in our experiments.  

\subsection{Implementation Detail}
The classification and detection models evaluated in this work are implemented in PyTorch \cite{paszke2019pytorch}. In the classification experiments, we use the Adam \cite{kingma2015adam} optimiser with the learning rates of $0.0001$ and $0.00001$ for the Bottle and Handgun datasets respectively.
In the 3D detection experiments, our models are also optimised by the Adam with a learning rate of $1e-4$ and stop training at 150 iterations. All experiments are conducted on a GTX 1080Ti GPU.

\section{Evaluation}\label{sec:evaluation}
We present and evaluate experimental results of both classification and detection in this section.
\subsection{Evaluation Criteria} \label{ssc:eval_criteria}
For the classification task, our model performances are evaluated in terms of True Positive rate (TPR\%) and False Positive rate (FPR\%). The mean and standard deviations over 10-fold cross validation are reported as the experimental results. For the detection task, we set the IoU (Intersection over Union; of ground truth and predicted 3D bounding boxes) threshold as 0.1 since it is significantly more challenging to get good overlapping bounding boxes in 3D object detection than that in 2D detection where an IoU threshold greater than 0.5 is typically used \cite{gaus2019evaluating}. Precision (\%) and Recall (\%) are calculated by thresholding the associated classification score at 0.9. In addition, Average Precision (\%) is reported to evaluate the overall model performance. All these evaluation criteria are reported as mean $\pm$ standard deviation over the three random splits of each dataset.

\subsection{3D Classification}
3D ResNet and Voxception-ResNet (VRN) are evaluated for 3D prohibited item classification in our experiments with evaluation results shown in Tables \ref{Table:cls_bot} and \ref{Table:cls_gun}. 

We firstly investigate the effect of different data pre-processing and augmentation strategies. For each model, the performance is reported when different combinations of strategies are used. It is obvious that \textit{rescaling} with $s=32$ is important to good classification performance. Without \textit{rescaling} the TPR is low and the FPR is high for ResNet$_{10}$ and ResNet$_{18}$ models on the \textit{Bottle} dataset and in the rest cases all models can not even converge (denoted by n/a in Tables \ref{Table:cls_bot}-\ref{Table:cls_gun}). The use of \textit{rescaling} significantly improves performance in all situations. Recall that \textit{rescaling} tends to reduce the difference of three dimensions (i.e., height, width and depth) of the input CT volumes, it is crucial to ensure the input volumes to have similar dimensions and suitable sizes so that the information loss can be avoided in the adaptive pooling layer.

The use of \textit{rich features} benefits the ResNet models on both datasets with increased TPR and reduced FPR consistently (Tables \ref{Table:cls_bot}, \ref{Table:cls_gun}). However, rich features do not make a difference for the VRN model. The reason is the VRN model has already employed the Inception-style architecture which is exactly for the fusion of multiple features learned with different kernel sizes. These results demonstrate the fact that rich features characterising information underlying different spatial scales are crucial for good classification performance and it can be implemented in different ways (e.g., multi-level feature volume fusion and inception-style network module).

The effect of \textit{rotation} varies on two datasets. For the \textit{Bottle} dataset, the use of \textit{rotation} does not improve the performance of any model (Table \ref{Table:cls_bot}). However, the data augmentation with \textit{rotation} benefits the classification of handguns significantly, especially for the ResNet models (Table \ref{Table:cls_gun}). One potential explanation is that the rotation operation applied into handgun signatures can generate more diversity than that into bottles.

By comparing different models in Tables \ref{Table:cls_bot},\ref{Table:cls_gun}, the VRN model achieves the highest TPR (98.9\% on bottles and 97.5\% on handguns) with moderate FPR (1.4\% on bottles and 1.5\% on handguns). As for the ResNet models, those having more layers generally perform better than the shallow ones. However, our results of ResNet$_{50}$ and ResNet$_{101}$ (not shown in tables) for classification demonstrate that deeper models are more difficult to converge with a limited number of training samples.

Table \ref{Table:comp} shows the comparison results of classification with prior work \cite{flitton20123d,mouton20143d,flitton2015object}. Overall, the best performing 3D CNN model VRN can achieve better results than the earlier Cortex \cite{flitton20123d} and Codebook \cite{flitton2015object} methods but is slightly worse than ERC \cite{mouton20143d} which, however, suffers from heavy computation burdens for computing dense 3D visual descriptors (187s per volume \cite{mouton20143d}). By contrast, our 3D CNN models are more efficient in the inference stage especially via parallelised GPU computation (0.0142s per volume).

\begin{table}[ht]
\centering
\caption{Classification results of varying 3D CNN architectures on the \textbf{Bottle} dataset.}\label{Table:cls_bot}
\begin{tabular}{lcccccc}
\hline
 \multirow{2}{*}{Model} &  \multicolumn{3}{c}{Augmentation} & \multirow{2}{*}{TPR (\%)} & \multirow{2}{*}{FPR (\%)} \\ \cline{2-4}
 &  Res & Rot & RF & & &  \\  \hline \hline
 \multirow{4}{*}{ ResNet$_{10}$}  & \xmark & \xmark & \xmark & 57.1 $\pm$ 16.2 & 26.7 $\pm$ 7.4 \\  
  & \cmark & \xmark & \xmark & 93.3 $\pm$ 3.5 & 1.6 $\pm$ 1.0 \\  
  & \cmark & \xmark & \cmark & 95.4 $\pm$ 2.9 & 0.9 $\pm$ 0.9 \\  
  & \cmark & \cmark & \cmark & 95.4 $\pm$ 4.0 & 0.9 $\pm$ 0.8 \\  
  \hline
 \multirow{4}{*}{ ResNet$_{18}$}  & \xmark & \xmark & \xmark & 61.3 $\pm$ 12.8 & 26.5 $\pm$ 22.6 \\  
  & \cmark & \xmark & \xmark & 94.7 $\pm$ 4.3 & \bf0.4 $\pm$ 0.8 \\  
  & \cmark & \xmark & \cmark & 94.8 $\pm$ 2.4 & 1.1 $\pm$ 2.5 \\  
  & \cmark & \cmark & \cmark & 96.0 $\pm$ 3.9 & 0.8 $\pm$ 0.8 \\  
  \hline
 \multirow{4}{*}{ ResNet$_{34}$}  & \xmark & \xmark & \xmark & n/a  & n/a  \\  
  & \cmark & \xmark & \xmark & 93.3 $\pm$ 6.7 & 2.9 $\pm$ 5.7 \\  
  & \cmark & \xmark & \cmark & 94.9 $\pm$ 3.5 & 0.7 $\pm$ 0.9 \\  
  & \cmark & \cmark & \cmark & 94.9 $\pm$ 2.6 & 0.8 $\pm$ 1.1 \\  
  \hline
  \multirow{4}{*}{Voxception-ResNet} & \xmark & \xmark & \xmark & 86.1 $\pm$ 10.4 & 14.7 $\pm$ 11.7 \\ 
  & \cmark & \xmark & \xmark & 98.9 $\pm$ 1.6 & 0.6 $\pm$ 0.7 \\ 
  & \cmark & \xmark & \cmark & 97.7 $\pm$ 2.4 & 0.8 $\pm$ 0.7 \\ 
  & \cmark & \cmark & \xmark & \bf98.9 $\pm$ 1.0 & 1.4 $\pm$ 1.1  \\ 
\hline
\end{tabular}
\end{table}

\begin{table}[ht]
\centering
\caption{Classification results of varying 3D CNN architectures on the \textbf{Handgun} dataset.}\label{Table:cls_gun}
\begin{tabular}{lcccccc}
\hline
 \multirow{2}{*}{Model} &  \multicolumn{3}{c}{Augmentation} & \multirow{2}{*}{TPR (\%)} & \multirow{2}{*}{FPR (\%)} \\ \cline{2-4}
 &  Res & Rot & RF & & &  \\  \hline \hline
\multirow{4}{*}{ ResNet$_{10}$}  & \xmark & \xmark & \xmark & n/a & n/a \\  
  & \cmark & \xmark & \xmark & 80.5 $\pm$ 6.5 & 10.8 $\pm$ 1.1 \\  
  & \cmark & \xmark & \cmark & 83.7 $\pm$ 6.0 & 8.3 $\pm$ 2.9 \\  
  & \cmark & \cmark & \cmark & 84.9 $\pm$ 9.5 & 11.3 $\pm$ 4.1 \\  
  \hline
 \multirow{4}{*}{ ResNet$_{18}$}  & \xmark & \xmark & \xmark & n/a & n/a \\  
  & \cmark & \xmark & \xmark & 82.6 $\pm$ 10.9 & 10.3 $\pm$ 1.6 \\  
  & \cmark & \xmark & \cmark & 85.1 $\pm$ 8.1 & 8.8 $\pm$ 3.8 \\  
  & \cmark & \cmark & \cmark & 89.4 $\pm$ 9.1 & 9.6 $\pm$ 5.2 \\  
  \hline
 \multirow{4}{*}{ ResNet$_{34}$}  & \xmark & \xmark & \xmark & n/a & n/a\\  
  & \cmark & \xmark & \xmark & 85.5 $\pm$ 6.0 & 7.6 $\pm$ 4.9 \\  
  & \cmark & \xmark & \cmark & 89.7 $\pm$ 5.5 & \bf0.8 $\pm$ 0.8 \\  
  & \cmark & \cmark & \cmark & 93.4 $\pm$ 4.6 & 3.6 $\pm$ 2.2\\  
  \hline
  \multirow{4}{*}{Voxception-ResNet} & \xmark & \xmark & \xmark & n/a & n/a \\ 
  & \cmark & \xmark & \xmark & 96.1 $\pm$ 4.9 & 1.5 $\pm$ 2.1 \\ 
  & \cmark & \xmark & \cmark & 94.7 $\pm$ 5.4 & 1.0 $\pm$ 1.0 \\ 
  & \cmark & \cmark & \xmark & \bf97.5 $\pm$ 2.4 & 1.5 $\pm$ 1.1  \\ 
\hline
\end{tabular}
\end{table}

\begin{table}[ht]
\centering
\caption{Comparison results of classification with prior work.}\label{Table:comp}
\begin{tabular}{lcccc}
\hline
 \multirow{2}{*}{Method} &  \multicolumn{2}{c}{Bottles} & \multicolumn{2}{c}{Guns}\\ \cline{2-5}
 & TPR (\%) & FPR (\%)& TPR (\%) & FPR (\%) \\  \hline \hline
Cortex \cite{flitton20123d} & 96.6 $\pm$ 3.2 & 1.0 $\pm$ 1.6 & 96.8 $\pm$ 2.6 & 1.1 $\pm$ 0.9\\
Codebook \cite{flitton2015object} & 89.3 $\pm$ 5.5 & 3.0 $\pm$ 1.4 & 97.3 $\pm$ 3.4 & 1.8 $\pm$ 1.7 \\
ERC \cite{mouton20143d} & \bf98.9 $\pm$ 0.7 & \bf0.6 $\pm$ 0.3 & \bf99.7 $\pm$ 0.5 & \bf0.3 $\pm$ 0.2 \\
VRN (Ours)  & \bf98.9 $\pm$ 1.0 & 1.4 $\pm$ 1.1 & 97.5 $\pm$ 2.4 & 1.5 $\pm$ 1.1  \\ 
\hline
\end{tabular}
\end{table}

\subsection{3D Object Detection} \label{sec:detResults}
\begin{table*}[ht]
	\centering
	\caption{Detection results of varying 3D CNN architectures on the \textbf{Bottle} dataset.}
	\label{table:det_bottle}
	\begin{tabular}{lllccc}
		\hline
		\multirow{2}{*}{Model} &  \multirow{2}{*}{Network} & 
		\multirow{2}{*}{Anchor size} & \multicolumn{2}{c}{Score threshold=0.9} & \multirow{2}{*}{Average Precision (\%)} \\ \cline{4-5}
		&  & & Precision (\%) & Recall (\%) & \\ \hline \hline 
		\multirow{8}{*}{Faster R-CNN \cite{ren2015fasterrcnn}}  & \multirow{4}{*}{ResNet$_{50}$} & 4-8-16-32 &  80.41 $\pm$ 3.77 & 62.47 $\pm$ 5.47 & 58.84 $\pm$ 5.91 \\
		& & 6-12-24-48 & 85.99 $\pm$ 2.82 & 68.56 $\pm$ 2.33 & 65.82 $\pm$ 3.13 \\
		& & 8-12-16-24 & 85.83 $\pm$ 2.81 & 65.66 $\pm$ 2.52 & 64.00 $\pm$ 3.10 \\
		& & 8-16-32-64 & \bf89.96 $\pm$ 3.65 & 67.56 $\pm$ 1.19 & 65.73 $\pm$ 0.73 \\ \cline{2-6}
		& \multirow{4}{*}{ResNet$_{101}$} & 4-8-16-32 & 74.79 $\pm$ 5.64 & 60.59 $\pm$ 9.34 & 54.34 $\pm$ 12.28 \\
		& & 6-12-24-48 & 87.18 $\pm$ 3.76 & 69.34 $\pm$ 1.26 & 66.74 $\pm$ 0.73 \\
		& & 8-12-16-24 & 84.40 $\pm$ 2.74 & 68.51 $\pm$ 2.89 & 65.95 $\pm$ 3.01 \\
		& & 8-16-32-64 & 83.28 $\pm$ 4.04 & 67.90 $\pm$ 1.89 & 64.77 $\pm$ 2.46 \\ \hline
		\multirow{8}{*}{RetinaNet \cite{lin2017focal}} & \multirow{4}{*}{ResNet$_{50}$} & 4-8-16-32 & 73.06 $\pm$ 10.05 & 74.46 $\pm$ 4.75 & 67.66 $\pm$ 9.86  \\
		& & 6-12-24-48 & 74.05 $\pm$ 2.49 & 81.71 $\pm$ 1.52 & 76.47 $\pm$ 2.76 \\
		& & 8-12-16-24 & 78.86 $\pm$ 4.59 & 81.00 $\pm$ 1.23 & 75.09 $\pm$ 2.10 \\
		& & 8-16-32-64 & 80.26 $\pm$ 5.75 & 78.30 $\pm$ 3.56 & 71.37 $\pm$ 1.52 \\ \cline{2-6} 
		
		& \multirow{4}{*}{ResNet$_{101}$} & 4-8-16-32 & 64.00 $\pm$ 6.44 & 67.77 $\pm$ 12.51 & 55.93 $\pm$ 14.06 \\
		& & 6-12-24-48 & 79.75 $\pm$ 3.75 & 78.74 $\pm$ 3.19 & 72.78 $\pm$ 3.32 \\
		& & 8-12-16-24 & 78.16 $\pm$ 1.89 & 80.14 $\pm$ 1.18 & 75.13 $\pm$ 1.21 \\
		& & 8-16-32-64 & 78.68 $\pm$ 2.16 & \bf82.42 $\pm$ 0.52 & \bf76.83 $\pm$ 1.09 \\ \hline
		
	\end{tabular}
\end{table*}

\begin{table*}[ht]
	\centering
	\caption{Detection results of varying 3D CNN architectures on the \textbf{Handgun} dataset.}
	\label{table:det_gun}
	\begin{tabular}{lllccc}
		\hline
		\multirow{2}{*}{Model} &  \multirow{2}{*}{Network} & 
		\multirow{2}{*}{Anchor size} & \multicolumn{2}{c}{Score threshold=0.9} & \multirow{2}{*}{Average Precision (\%)} \\ \cline{4-5}
		&  & & Precision (\%) & Recall (\%) & \\ \hline \hline 
		\multirow{8}{*}{Faster R-CNN \cite{ren2015fasterrcnn}}  & \multirow{4}{*}{ResNet$_{50}$} & 4-8-16-32 &   91.67 $\pm$ 1.82 & 84.62 $\pm$ 2.21 & 84.00 $\pm$ 2.07 \\
		& & 6-12-24-48 & 91.38 $\pm$ 4.19 & 86.98 $\pm$ 2.25 & 85.30 $\pm$ 3.11 \\
		& & 8-12-16-24 & \bf 92.43 $\pm$ 1.37 & 86.38 $\pm$ 1.74 & 85.40 $\pm$ 1.77 \\
		& & 8-16-32-64 & 91.98 $\pm$ 3.33 & 87.56 $\pm$ 1.54 & 86.74 $\pm$ 1.81 \\ \cline{2-6}
		& \multirow{4}{*}{ResNet$_{101}$} & 4-8-16-32 & 89.93 $\pm$ 3.39 & 85.18 $\pm$ 4.74 & 83.98 $\pm$ 5.03 \\
		& & 6-12-24-48 & 91.00 $\pm$ 4.23 & 88.74 $\pm$ 1.76 & 87.76 $\pm$ 1.82 \\
		& & 8-12-16-24 & 91.70 $\pm$ 1.07 & 85.19 $\pm$ 2.31 & 84.38 $\pm$ 2.16 \\
		& & 8-16-32-64 & 90.17 $\pm$ 1.76 & 86.97 $\pm$ 3.41 & 85.92 $\pm$ 2.93 \\ \hline
		\multirow{8}{*}{RetinaNet \cite{lin2017focal}} & \multirow{4}{*}{ResNet$_{50}$} &  4-8-16-32 &  87.29 $\pm$ 2.14 & 89.34 $\pm$ 1.52 & 87.30 $\pm$ 3.33 \\
		& & 6-12-24-48 & 90.06 $\pm$ 2.79 & 90.53 $\pm$ 0.88 & \bf89.13 $\pm$ 0.87 \\
		& & 8-12-16-24 & 91.15 $\pm$ 3.63 & 89.95 $\pm$ 1.58 & 88.12 $\pm$ 2.33 \\
		& & 8-16-32-64 & 88.98 $\pm$ 2.67 & 89.94 $\pm$ 0.91 & 85.89 $\pm$ 1.74 \\ \cline{2-6} 
		& \multirow{4}{*}{ResNet$_{101}$} & 4-8-16-32 & 88.95 $\pm$ 2.21 & 90.53 $\pm$ 2.24 & 88.61 $\pm$ 2.25 \\
		& & 6-12-24-48 & 89.69 $\pm$ 4.30 & \bf 90.55 $\pm$ 2.13 & 87.82 $\pm$ 2.65 \\
		& & 8-12-16-24 & 91.09 $\pm$ 1.33 & 90.54 $\pm$ 0.75 & 87.31 $\pm$ 1.35 \\
		& & 8-16-32-64 & 90.56 $\pm$ 2.10 & 90.54 $\pm$ 1.64 & 87.18 $\pm$ 2.40 \\ \hline
		
	\end{tabular}
\end{table*}


For prohibited item (i.e., bottles and handguns) detection within 3D X-ray CT baggage screening images, we employ Faster R-CNN \cite{ren2015fasterrcnn} and RetinaNet \cite{lin2017focal} CNN architectures as set out in Section \ref{sec:method_det}. The detection results using ResNet$_{50}$ and ResNet$_{101}$ backbone netwroks are presented in the Tables \ref{table:det_bottle} and \ref{table:det_gun}. 

In our detection experiments we investigate the effect of different factors, such as {\it resampling}, {\it anchor box size}, and {\it confidence threshold}. We report the best performing combinations by varying configurations for both detection models. The {resampling} is essential to achieve good detection performance. We observe, for both detection architectures on both datasets, {resampling} CT volume by $1/3$ in all three dimensions significantly achieve better results (AP: 64.77 {\it Bottle} dataset, Fatser R-CNN with ResNet$_{101}$) compared to resampling factor of $1/2$ (AP: 58.68 {\it Bottle} dataset, Fatser R-CNN with ResNet$_{101}$). Therefore, the results reported in the Tables \ref{table:det_bottle} and \ref{table:det_gun}, the resampling factor of $1/3$ is applied with confidence score threshold of 0.9.

We vary the anchor sizes ($4$ different sets) in our experiments as explained in the Section \ref{sec:method_det}. The AP is higher with the larger anchor size, i.e., (\textit{6-12-24-48}), (\textit{8-16-32-64}) (AP: $\sim$65\%), compared to anchor size of (\textit{4-8-16-32}) (AP: $\sim$58\%) while using Faster R-CNN \cite{ren2015fasterrcnn} with ResNet$_{50}$ (Table \ref{table:det_bottle} upper) for {\it Bottle} dataset. The similar trend is perceptible for both Faster R-CNN \cite{ren2015fasterrcnn} and RetinaNet \cite{lin2017focal} (Table \ref{table:det_bottle}, upper and lower) with ResNet$_{50}$ and ResNet$_{101}$. The best AP is achieved by RetinaNet with ResNet$_{101}$ (AP: 76\%, Table \ref{table:det_bottle}, lower) with anchor size of (\textit{8-16-32-64}). For {\it Handgun} dataset, RetinaNet with ResNet$_{50}$ achieves the highest average precision (AP: 89\%, Table \ref{table:det_gun}, lower) using (\textit{6-12-24-18}) anchor size. It observable that for {\it Handgun} dataset, all different variant of anchor sizes achieve similar performances on all the metrics (precision, recall and AP).   
 
 From the results (Tables \ref{table:det_bottle}, \ref{table:det_gun}), by increasing the number of convolutional layers in backbone network (ResNet$_{50}$ vs ResNet$_{101}$) does not increase the performance. By comparing two different detection architectures, RetinaNet \cite{lin2017focal} outperforms Faster R-CNN \cite{ren2015fasterrcnn} for both {\it Botte} dataset (AP: 76\%, Table \ref{table:det_bottle}, lower) and {\it Handgun} dataset (AP: 89\%, Table \ref{table:det_gun}, lower). 
 
\begin{figure*}
    \centering
    {\includegraphics[width=0.88\textwidth]{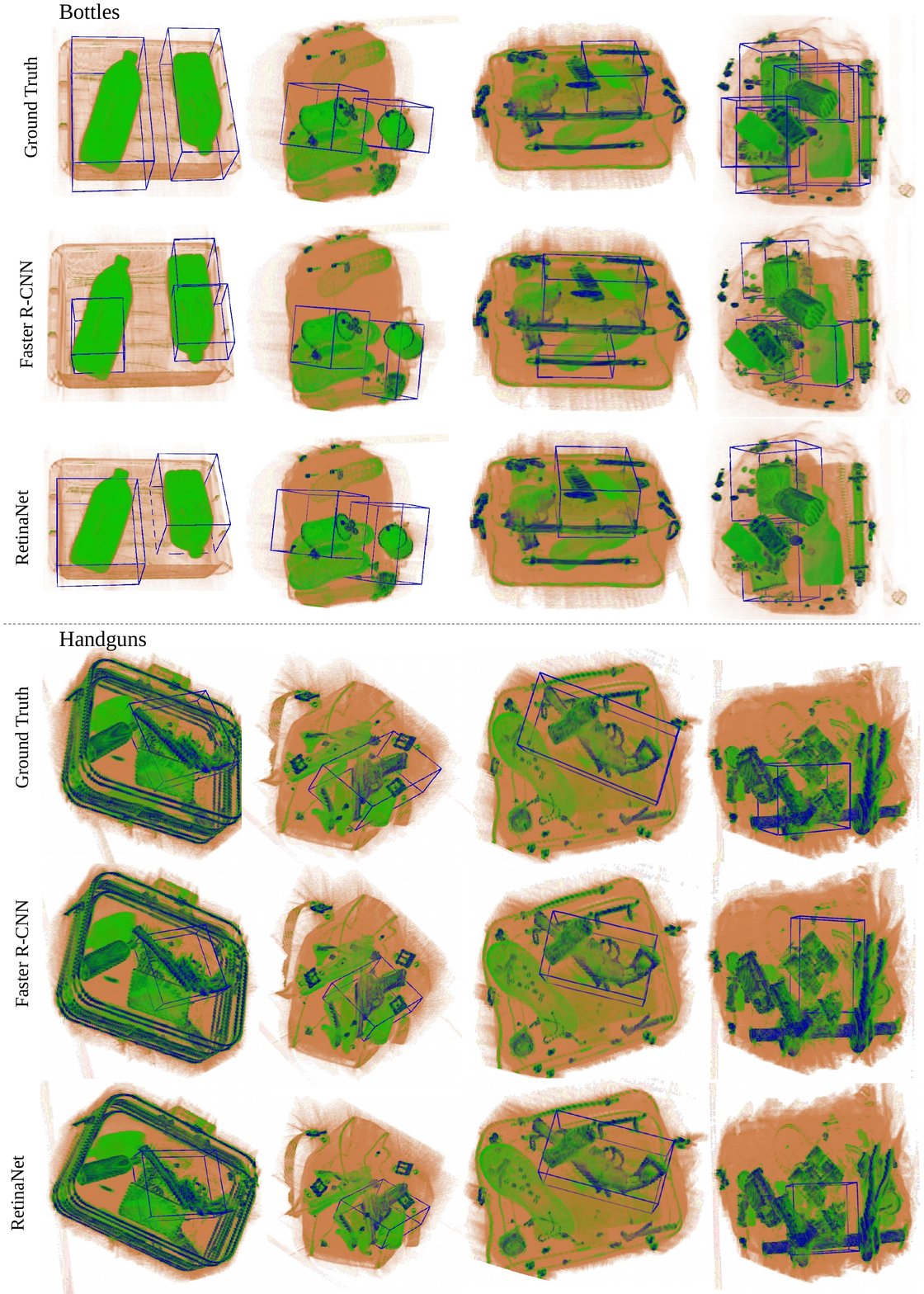}}
    {\caption{Exemplar detection results (ground truth, detection results of Faster R-CNN \cite{ren2015fasterrcnn} and RetinaNet \cite{lin2017focal} both with ResNet$_{101}$ are shown in top three rows for bottles and bottom three rows for handguns).}
    \label{fig:det_results}}
\end{figure*}

 Exemplar prohibited items detection results from Faster R-CNN \cite{ren2015fasterrcnn} and RetinaNet \cite{lin2017focal} with ResNet$_{101}$ are depicted in Figure \ref{fig:det_results}. 
 From the examples, we observe Faster R-CNN \cite{ren2015fasterrcnn} falsely detects a bottle while RetinaNet \cite{lin2017focal} correctly detects the item (Figure \ref{fig:det_results}-Bottles, column 3). This is anticipated due to the superior performance of RetinaNet \cite{lin2017focal} than Faster R-CNN \cite{ren2015fasterrcnn} for {\it Bottle} dataset (Table \ref{Table:cls_bot}). 
 Both the models perform in a similar fashion for handguns as depicted in the Figure \ref{fig:det_results}-Handguns, echoed our quantitative evaluations in Table \ref{Table:cls_gun}.      
 
\section{Conclusion} \label{sec:conclusion}
We extend Convolutional Neural Networks for prohibited item classification and detection in volumetric 3D CT baggage security screening X-ray imagery. 
As the first attempt of deep CNN based techniques in this specific application, we make extensive evaluations on a variety of CNN models and data pre-processing strategies. The experimental results on classification demonstrate comparable performance (TP: $\sim$98\%, FP: $\sim$1.5\% for both {\it Bottle} and {\it Handgun}) of the Voxception-ResNet model with prior art using hand-crafted 3D features whilst the 3D CNN model is more computationally efficient than the traditional methods \cite{mouton20143d}. The results of detection demonstrate the feasibility of extending traditional 2D object detectors (e.g., Faster R-CNN and RetinaNet) to detect prohibited item in volumetric 3D CT data (mAP: $\sim$76\% for {\it Bottle}, mAP: $\sim$88\% for {\it Handgun}).

Although viable performance has been achieved for the classification task, the detection task still needs to be improved in order to meet operational requirements for aviation security screening. Limited training data is the primary reason for the more limited detection performance. 
In our future work, we will take advantage of transfer learning and the use of synthetic data \cite{wang2020reference} to improve the performance. In addition, we will incorporate variety of prohibited items for multi-class classification and detection problems.

\ifCLASSOPTIONcaptionsoff
  \newpage
\fi

\bibliographystyle{IEEEtran}
\bibliography{ref}

\begin{thebibliography}{10}
\providecommand{\url}[1]{#1}
\csname url@samestyle\endcsname
\providecommand{\newblock}{\relax}
\providecommand{\bibinfo}[2]{#2}
\providecommand{\BIBentrySTDinterwordspacing}{\spaceskip=0pt\relax}
\providecommand{\BIBentryALTinterwordstretchfactor}{4}
\providecommand{\BIBentryALTinterwordspacing}{\spaceskip=\fontdimen2\font plus
\BIBentryALTinterwordstretchfactor\fontdimen3\font minus
  \fontdimen4\font\relax}
\providecommand{\BIBforeignlanguage}[2]{{%
\expandafter\ifx\csname l@#1\endcsname\relax
\typeout{** WARNING: IEEEtran.bst: No hyphenation pattern has been}%
\typeout{** loaded for the language `#1'. Using the pattern for}%
\typeout{** the default language instead.}%
\else
\language=\csname l@#1\endcsname
\fi
#2}}
\providecommand{\BIBdecl}{\relax}
\BIBdecl

\bibitem{bhowmik2019good}
N.~Bhowmik, Q.~Wang, Y.~F.~A. Gaus, M.~Szarek, and T.~P. Breckon, ``The good,
  the bad and the ugly: Evaluating convolutional neural networks for prohibited
  item detection using real and synthetically composited {X}-ray imagery,'' in
  \emph{British Machine Vision Conference Workshops}, 2019.

\bibitem{wang2020reference}
Q.~Wang, N.~Megherbi, and T.~P. Breckon, ``A reference architecture for
  plausible threat image projection ({TIP}) within 3{D} {X}-ray computed
  tomography volumes,'' \emph{Journal of X-ray Science and Technology}, 2020,
  in press.

\bibitem{meuter2016and}
R.~F. Meuter and P.~F. Lacherez, ``When and why threats go undetected: Impacts
  of event rate and shift length on threat detection accuracy during airport
  baggage screening,'' \emph{Human factors}, vol.~58, no.~2, pp. 218--228,
  2016.

\bibitem{akcay2018using}
S.~Akcay, M.~E. Kundegorski, C.~G. Willcocks, and T.~P. Breckon, ``Using deep
  convolutional neural network architectures for object classification and
  detection within x-ray baggage security imagery,'' \emph{IEEE transactions on
  information forensics and security}, vol.~13, no.~9, pp. 2203--2215, 2018.

\bibitem{gaus2019evaluating}
Y.~Gaus, N.~Bhowmik, S.~Akcay, and T.~Breckon, ``Evaluating the transferability
  and adversarial discrimination of convolutional neural networks for threat
  object detection and classification within x-ray security imagery,'' in
  \emph{Proc. Int. Conf. on Machine Learning Applications}.\hskip 1em plus
  0.5em minus 0.4em\relax IEEE, December 2019.

\bibitem{bhowmik19subcomponent}
N.~Bhowmik, Y.~Gaus, S.~Akcay, J.~Barker, and T.~Breckon, ``On the impact of
  object and sub-component level segmentation strategies for supervised anomaly
  detection within x-ray security imagery,'' in \emph{Proc. Int. Conf. on
  Machine Learning Applications}.\hskip 1em plus 0.5em minus 0.4em\relax IEEE,
  December 2019, to appear.

\bibitem{bhowmik19electronics}
N.~Bhowmik, Y.~Gaus, and T.~Breckon, ``Using deep neural networks to address
  the evolving challenges of concealed threat detection within complex
  electronic items,'' in \emph{Proc. Conference on Homeland Security}.\hskip
  1em plus 0.5em minus 0.4em\relax IEEE, November 2019, to appear.

\bibitem{he2016deep}
K.~He, X.~Zhang, S.~Ren, and J.~Sun, ``Deep residual learning for image
  recognition,'' in \emph{Proc. computer vision and pattern recognition}, 2016,
  pp. 770--778.

\bibitem{brock2016generative}
A.~Brock, T.~Lim, J.~M. Ritchie, and N.~Weston, ``Generative and discriminative
  voxel modeling with convolutional neural networks,'' in \emph{Neural
  Information Processing Systems}, 2016.

\bibitem{ren2015fasterrcnn}
S.~Ren, K.~He, R.~Girshick, and J.~Sun, ``Faster r-cnn: Towards real-time
  object detection with region proposal networks,'' in \emph{Advances in neural
  information processing systems}, 2015, pp. 91--99.

\bibitem{lin2017focal}
T.-Y. Lin, P.~Goyal, R.~Girshick, K.~He, and P.~Doll{\'a}r, ``Focal loss for
  dense object detection,'' in \emph{Proc. Int. conf. on computer vision},
  2017, pp. 2980--2988.

\bibitem{wiley2012automatic}
D.~F. Wiley, D.~Ghosh, and C.~Woodhouse, ``Automatic segmentation of {CT} scans
  of checked baggage,'' in \emph{Proc. Int. Meeting on Image Formation in X-ray
  {CT}}, 2012, pp. 310--313.

\bibitem{flitton20123d}
G.~Flitton, T.~P. Breckon, and N.~Megherbi, ``A 3{D} extension to cortex like
  mechanisms for 3{D} object class recognition,'' in \emph{Proc. Computer
  Vision and Pattern Recognition}.\hskip 1em plus 0.5em minus 0.4em\relax IEEE,
  2012, pp. 3634--3641.

\bibitem{mouton20143d}
A.~Mouton, T.~P. Breckon, G.~T. Flitton, and N.~Megherbi, ``3{D} object
  classification in baggage computed tomography imagery using randomised
  clustering forests,'' in \emph{Proc. Int. conf. on image processing (ICIP)},
  2014, pp. 5202--5206.

\bibitem{jin2015joint}
P.~Jin, D.~H. Ye, and C.~A. Bouman, ``Joint metal artifact reduction and
  segmentation of {CT} images using dictionary-based image prior and
  continuous-relaxed potts model,'' in \emph{Proc. Int. conf. on Image
  Processing (ICIP)}.\hskip 1em plus 0.5em minus 0.4em\relax IEEE, 2015, pp.
  798--802.

\bibitem{flitton2015object}
G.~Flitton, A.~Mouton, and T.~P. Breckon, ``Object classification in 3{D}
  baggage security computed tomography imagery using visual codebooks,''
  \emph{Pattern Recognition}, vol.~48, no.~8, pp. 2489--2499, 2015.

\bibitem{mouton2015materials}
A.~Mouton and T.~P. Breckon, ``Materials-based 3{D} segmentation of unknown
  objects from dual-energy computed tomography imagery in baggage security
  screening,'' \emph{Pattern Recognition}, vol.~48, no.~6, pp. 1961--1978,
  2015.

\bibitem{mouton2015review}
------, ``A review of automated image understanding within 3{D} baggage
  computed tomography security screening,'' \emph{Journal of X-ray Science and
  Technology}, vol.~23, no.~5, pp. 531--555, 2015.

\bibitem{wang2019approach}
Q.~Wang, K.~N. Ismail, and T.~P. Breckon, ``An approach for adaptive automatic
  threat recognition within 3{D} computed tomography images for baggage
  security screening,'' \emph{Journal of X-ray Science and Technology}, 2019.

\bibitem{maturana2015voxnet}
D.~Maturana and S.~Scherer, ``Voxnet: A 3{D} convolutional neural network for
  real-time object recognition,'' in \emph{Proc. Int. conf. on Intelligent
  Robots and Systems}.\hskip 1em plus 0.5em minus 0.4em\relax IEEE, 2015, pp.
  922--928.

\bibitem{zhou2018voxelnet}
Y.~Zhou and O.~Tuzel, ``Voxelnet: End-to-end learning for point cloud based
  3{D} object detection,'' in \emph{Proc. Computer Vision and Pattern
  Recognition}, 2018, pp. 4490--4499.

\bibitem{qi2018frustum}
C.~R. Qi, W.~Liu, C.~Wu, H.~Su, and L.~J. Guibas, ``Frustum pointnets for 3{D}
  object detection from rgb-d data,'' in \emph{Proc. Computer Vision and
  Pattern Recognition}, 2018, pp. 918--927.

\bibitem{xie2019automated}
H.~Xie, D.~Yang, N.~Sun, Z.~Chen, and Y.~Zhang, ``Automated pulmonary nodule
  detection in {CT} images using deep convolutional neural networks,''
  \emph{Pattern Recognition}, vol.~85, pp. 109--119, 2019.

\bibitem{jaeger2019retina}
P.~F. Jaeger, S.~A. Kohl, S.~Bickelhaupt, F.~Isensee, T.~A. Kuder, H.-P.
  Schlemmer, and K.~H. Maier-Hein, ``Retina u-net: Embarrassingly simple
  exploitation of segmentation supervision for medical object detection,'' in
  \emph{Proc. Neural Information Processing Systems Workshops}, 2019.

\bibitem{qi2016volumetric}
C.~R. Qi, H.~Su, M.~Nie{\ss}ner, A.~Dai, M.~Yan, and L.~J. Guibas, ``Volumetric
  and multi-view cnns for object classification on 3{D} data,'' in \emph{Proc.
  computer vision and pattern recognition}, 2016, pp. 5648--5656.

\bibitem{qi2017pointnet}
C.~R. Qi, H.~Su, K.~Mo, and L.~J. Guibas, ``Pointnet: Deep learning on point
  sets for 3{D} classification and segmentation,'' in \emph{Proc. Computer
  Vision and Pattern Recognition}, 2017, pp. 652--660.

\bibitem{chen2019med3d}
S.~Chen, K.~Ma, and Y.~Zheng, ``Med3{D}: Transfer learning for 3{D} medical
  image analysis,'' \emph{arXiv preprint arXiv:1904.00625}, 2019, unpublished.

\bibitem{girshick2014rich}
R.~Girshick, J.~Donahue, T.~Darrell, and J.~Malik, ``Rich feature hierarchies
  for accurate object detection and semantic segmentation,'' in \emph{Proc.
  computer vision and pattern recognition}, 2014, pp. 580--587.

\bibitem{szegedy2016rethinking}
C.~Szegedy, V.~Vanhoucke, S.~Ioffe, J.~Shlens, and Z.~Wojna, ``Rethinking the
  inception architecture for computer vision,'' in \emph{Proc. computer vision
  and pattern recognition}, 2016, pp. 2818--2826.

\bibitem{redmon2016you}
J.~Redmon, S.~Divvala, R.~Girshick, and A.~Farhadi, ``You only look once:
  Unified, real-time object detection,'' in \emph{Proc. computer vision and
  pattern recognition}, 2016, pp. 779--788.

\bibitem{paszke2019pytorch}
A.~Paszke, S.~Gross, F.~Massa, A.~Lerer, J.~Bradbury, G.~Chanan, T.~Killeen,
  Z.~Lin, N.~Gimelshein, L.~Antiga \emph{et~al.}, ``Pytorch: An imperative
  style, high-performance deep learning library,'' in \emph{Advances in Neural
  Information Processing Systems}, 2019, pp. 8024--8035.

\bibitem{kingma2015adam}
D.~P. Kingma and J.~Ba, ``Adam: A method for stochastic optimization,'' in
  \emph{Proc. Int. conf. on Learning Representations}, 2015.

\end{thebibliography}

\end{document}